\documentclass[sigconf]{acmart}
\AtBeginDocument{%
  \providecommand\BibTeX{{%
    \normalfont B\kern-0.5em{\scshape i\kern-0.25em b}\kern-0.8em\TeX}}}

\usepackage{amsmath,amssymb,amsfonts}
\usepackage{amsthm}
\usepackage{graphicx}
\usepackage{subfigure}
\usepackage{caption}
\setcopyright{acmlicensed}

\begin{document}

\title{Cross-Global Attention Graph Kernel Network Prediction of Drug Prescription}

\author{Hao-Ren Yao}
\email{hao-ren@ir.cs.georgetown.edu}
\affiliation{
  \institution{Georgetown University}
  \state{Washington, D.C.}
  \country{USA}
}

\author{Der-Chen Chang}
\email{chang@georgetown.edu}
\affiliation{
  \institution{Georgetown University}
  \state{Washington, D.C.}
  \country{USA}
}

\author{Ophir Frieder}
\email{ophir@ir.cs.georgetown.edu}
\affiliation{
  \institution{Georgetown University}
  \state{Washington, D.C.}
  \country{USA}
}

\author{Wendy Huang, MD}
\email{al0357186@hotmail.com}
\affiliation{
  \institution{Meng Cheng Family Medicine Clinic}
  \state{Kaohsiung City}
  \country{Taiwan}
}

\author{I-Chia Liang, MD}
\email{ysonyaliang@gmail.com}
\affiliation{
  \institution{Tri-Service General Hospital}
  \state{Taipei City}
  \country{Taiwan}
}

\author{Chi-Feng Hung}
\email{054317@gapp.fju.edu.tw}
\affiliation{%
  \institution{Fu Jen Catholic University}
  \state{New Taipei City}
  \country{Taiwan}
}

\renewcommand{\shortauthors}{H.-R. Yao et al.}

\copyrightyear{2020}
\acmYear{2020}
\setcopyright{acmlicensed}\acmConference[BCB '20]{Proceedings of the 11th ACM International Conference on Bioinformatics, Computational Biology and Health Informatics}{September 21--24, 2020}{Virtual Event, USA}
\acmBooktitle{Proceedings of the 11th ACM International Conference on Bioinformatics, Computational Biology and Health Informatics (BCB '20), September 21--24, 2020, Virtual Event, USA}
\acmPrice{15.00}
\acmDOI{10.1145/3388440.3412459}
\acmISBN{978-1-4503-7964-9/20/09}

\begin{abstract}
  We present an end-to-end, interpretable, deep-learning architecture to learn a graph kernel that predicts the outcome of chronic disease drug prescription.  This is achieved through a deep metric learning collaborative with a Support Vector Machine objective using a graphical representation of Electronic Health Records. We formulate the predictive model as a binary graph classification problem with an adaptive learned graph kernel through novel cross-global attention node matching between patient graphs, simultaneously computing on multiple graphs without training pair or triplet generation. Results using the Taiwanese National Health Insurance Research Database demonstrate that our approach outperforms current start-of-the-art models both in terms of accuracy and interpretability.
\end{abstract}

\ccsdesc[500]{Computing methodologies~Artificial intelligence}
\ccsdesc[500]{Applied computing~Health informatics}

\keywords{Health informatics; Predictive model; Deep Graph Kernel Learning; Graph kernel}
\maketitle

\section{Introduction}
\label{introduction}

Outcome prediction of chronic disease drug prescription is a preeminent yet unsolved problem. Chronic diseases are a major cause of illness in the United States\footnote{https://www.cdc.gov/chronicdisease/index.htm} and a top ten cause of death in Taiwan\footnote{https://www.mohw.gov.tw/cp-4650-50697-2.html}. Chronic disease drug prescription aims to reduce patient risk for severe comorbidities and complications. Prescribing such medication is difficult as long-term disease progression and numerous other factors complicate treatment plan design. On the other hand, the availability of Electronic Health Records (EHRs), providing historical medical road-maps for patients, enable the development of intelligent predictive systems for drug prescription ~\cite{PH1, jackson2011}.

Various EHR modelling approaches, including electronic phenotyping (e.g., feature extraction) ~\cite{acm-ehr-survey-2017} and highly-accurate deep learning models (e.g., representation learning) ~\cite{deep-patient}, support such analytical tasks.  For example, Recurrent Neural Networks (RNN) ~\cite{lstm} model time series medical data. However, 
interpretability concerns associated with deep learning approaches, 
particularly in the medical domain, limit their use. Notwithstanding, the trade-off to achieve high accuracy and high interpretability remains. 

Many studies introduce attention-based RNN models to improve interpretability ~\cite{retain, dipole}. However, the majority of efforts rely on publicly available datasets or on a collaborating hospital's EHR system where patient demographic information is mostly uniform. Unfortunately, this uniformity of data fails to exist when developing approaches for real-world, integrated EHR systems (e.g., insurance claim-based EHR systems). On this occasion, highly temporally dependent data attributes with high noise and variance often induce model over-fitting. Such a problem is addressed in ~\cite{yao2019graph, yao2019multiple} with a proposed graph-kernel EHR predictive model, yet they only consider a single medication with immediate outcome observations. For chronic diseases, long-term disease progression coupled with EHR complexity complicates the effort. We surmise that attention-based deep learning models and handcrafted kernel computations are limited to handle complex EHR under long-term disease progression. As discussed in ~\cite{yao2019multiple}, the increased divergence and noise on data attributes over-fits the deep learning model and defeats the handcrafted kernel.

We propose a cross-global attention graph kernel network to learn optimal graph kernels on a graphical representation of patient EHRs. We term "cross-global" to delineate pairwise-less "cross" graph node attention and its "global" attention graph pooling. The novel cross-global attention node matching automatically captures relevant information in biased long-term disease progression. In contrast to attention-based graph similarity learning ~\cite{bai2019simgnn, li2019graph, al2019ddgk} that relies on a pairwise comparisons of training pairs or triplets, our matching is performed on a batch of graphs simultaneously by a global cluster membership assignment. This is accomplished without the need to generate training pairs or triplets for pairwise computations and seamlessly combines classification loss. The learning process is guided by cosine distance. The resulting kernel, compared to its Euclidean distance counterpart, has better noise resistance under a high dimension space ~\cite{calin2009subriemannian, calin2010heat}. Unlike distance metric learning ~\cite{hadsell2006dimensionality, schroff2015facenet} and aforementioned graph similarity learning, we align our learned distance and graph kernel to a classification objective. We formulate an end to end training by jointly optimizing contrastive and kernel alignment loss with a Support Vector Machine (SVM) primal objective. Such a training procedure encourages node matching and similarity measurement to produce ideal classification, providing interpretation on prediction. The resulting kernel function can be directly used by an off-the-shelf kernelized classifier (e.g., scikit-learn SVC~\footnote{https://scikit-learn.org/stable/modules/svm.html}). The cross-global attention node matching and kernel-based classification makes it interpretable in both knowledge discovery and prediction case study. 

We evaluate our model using a country-wide population, claim-based database from Taiwan; the National Health Insurance Research Database (NHIRD). We formulate the chronic disease drug prediction task as a binary graph classification problem. An optimal graph kernel learned through cross-global attention graph kernel network is used to perform classification on a kernel SVM. Experimental results demonstrate that our proposed method outperforms current state-of-the-art approaches as well as providing model interpretability. Analysis on node matching between patient graphs indicates how our cluster membership assignment can generate effective node matching without explicit pairwise computation. We also demonstrate  superior interpretability over node matching on most similar cases and support vectors, serving as knowledge and information discovery on prediction. We are the first to combine pairwise-less graph kernel learning and classification objective in an end to end learning procedure for medical practice. Our approach is  under clinical used and evaluation.  

Our contributions are as follows:
\begin{itemize}
    \item We propose an end-to-end, deep metric learning based framework to learn an optimal graph kernel on highly noisy EHR data.
    \item We present a pairwise-less attention-based node matching operation and metric-learning process without the need to generate training pairs or triplets to perform pairwise similarity measurement, seamlessly combining SVM objectives.
    \item We experiment with large-scale, real-world, long-term span medical data to demonstrate our effectiveness together with interpretability, surpassing all state-of-the-art baselines.
    \item We provide a clinically-vetted approach.
\end{itemize}

\begin{figure}
\centering
\includegraphics[scale=0.12]{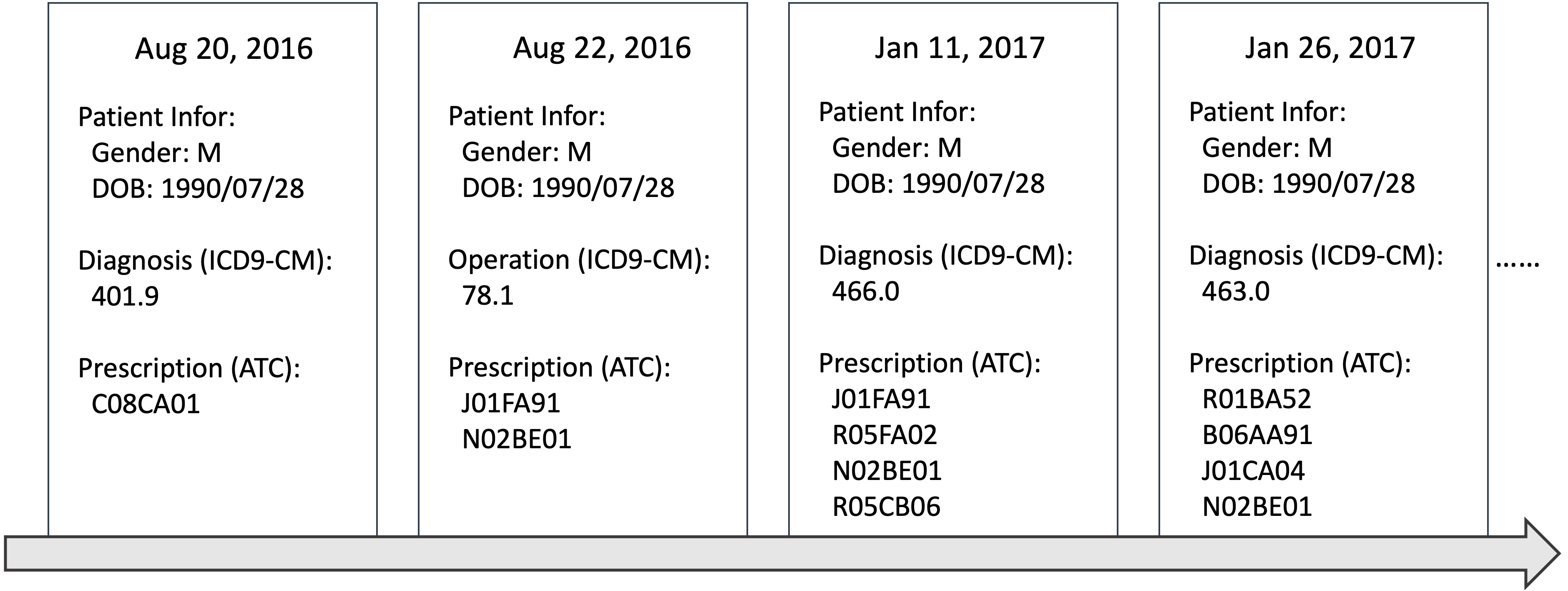}
\caption{A sample patient EHRs.}
\label{subset-ehr}
\end{figure}

\begin{figure}
\centering
\includegraphics[scale=0.08]{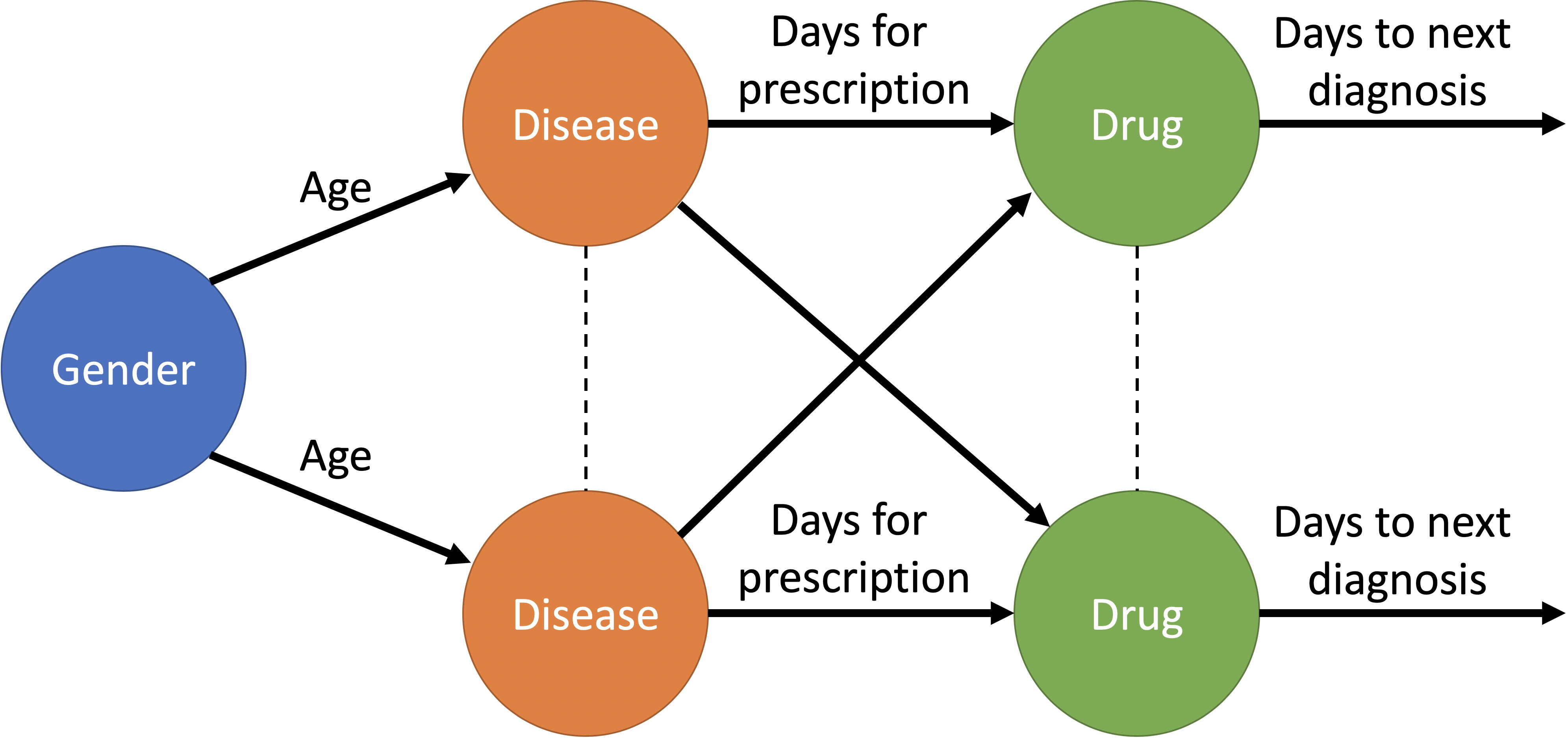}
\caption{An example of patient graph}
\label{patient-graph}
\end{figure}

\section{Related work}
Most drug prescription prediction tasks focus on Adverse Drug Reactions (ADRs) or medication errors ~\cite{ME1, yates2015AAAI, xiao2017adverse, nguyen2013probabilistic} while other efforts discuss the effectiveness of drug prescription for a given disease diagnosis ~\cite{jackson2011, kang2015efficient}. The approaches mainly discuss ADRs and specific disease target information and are unable to utilize EHR for outcome prediction and complex disease progression modeling from medical history. 

Despite traditional electronic phenotyping ~\cite{acm-ehr-survey-2017}, representation learning on EHR benefits from deep learning models. Most efforts use deep architectures to learn EHR embeddings via Multi-layer Perceptron (MLP) ~\cite{deep-patient, med2vec}, Convolutional Neural Network (CNN) ~\cite{CNN} or Recurrent Neural Network (RNN) to model time series medical information ~\cite{lstm}. An Attention-based model was also  proposed to address model interpretability ~\cite{retain, dipole, choi2017gram, mullenbach2018explainable, xie2019ehr}. Recently, BERT trained on clinical language (Clinical BERT) was introduced to support various fine-tuning tasks such as hospital readmission prediction ~\cite{huang2019clinicalbert, clinicalBert}. However, most efforts concentrate on medical code prediction and medical concept embedding and do not directly map onto a drug prediction task. Moreover, the model over-fitting and interpretability trade-off still remains unsolved. In ~\cite{yao2019graph, yao2019multiple}, a graph kernel approach is developed to predict outcomes of a drug prescription for a given disease diagnosis, which achieves state-of-the-art results. However, as mentioned in Section ~\ref{introduction} and evaluated in Section \ref{results}, chronic diseases are not considered in the model, leading to under-performance prediction. Overall, the EHR prediction task on drug prescription is not fully investigated. 

\section{Prediction Task on chronic disease drug prescription}
\subsection{EHR Patient Graph}
We formulate a patient's EHR as a Directed Acyclic Graph (DAG) following the definition in ~\cite{yao2019graph}, for which each node represents a medical event, and an edge between two nodes represents an ordering with the time difference as edge weight (e.g., days). The demographic information of the patient, e.g., gender, connects to the first medical event with age as an edge weight. Figure ~\ref{patient-graph} describes an example patient graph. As in ~\cite{yao2019graph}, we only use gender and age as demographic information to simplify the model. All node labels are one-hot encoded.

\subsection{Success and Failure cases}
\begin{figure}
\centering
\vfill
\subfigure[Success case]{\includegraphics[scale=0.15]{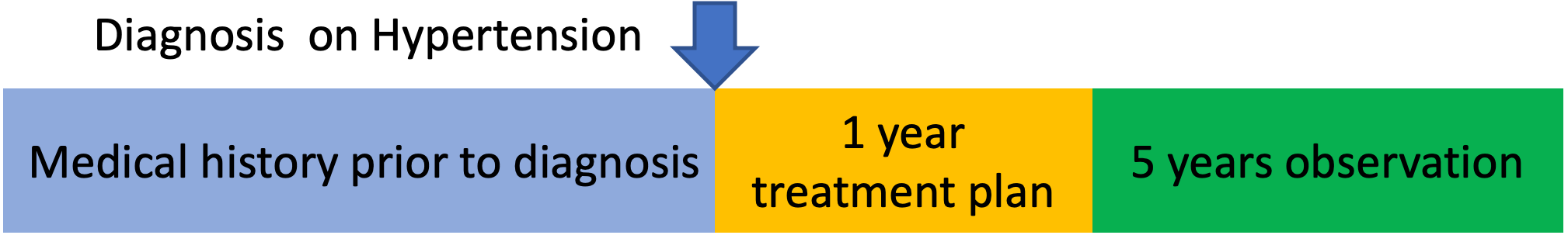}}
\vfill
\subfigure[Failure case]{\includegraphics[scale=0.15]{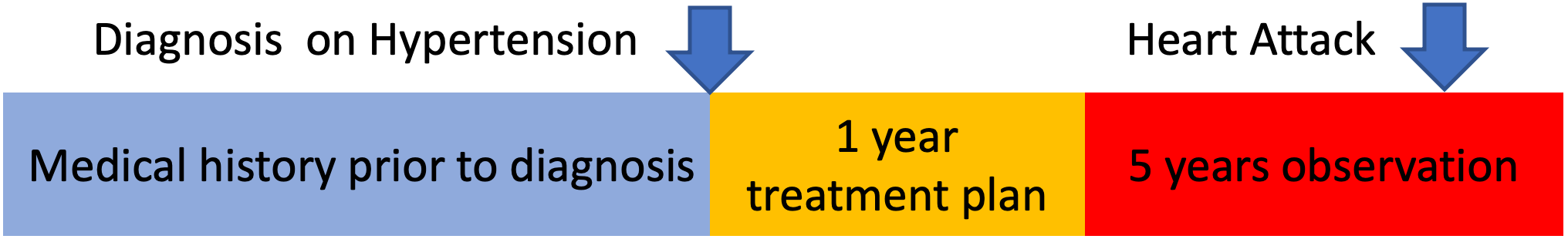}}
\caption{Criteria for success and failure cases}
\label{success-failure-case}
\end{figure}

To define the success or failure of a treatment plan~\footnote{Since chronic diseases require a set of drug prescriptions with necessary adjustment per disease condition, we use the term treatment plan and drug prescription interchangeably.} for a chronic disease, we follow the guideline published by the National Medical Association for selected chronic diseases ~\cite{chiang20152015,li20172017,diabetes-guideline}. Generally, an observation window is defined after a treatment period to monitor whether the given treatment plan achieves its treatment objective (e.g., no severe complication occurrence in 5 years). Given a chronic disease diagnosis, a treatment is considered a failure if the patient is diagnosed with a selected severe complication or comorbidity within the post treatment observation window. Otherwise, the treatment is considered a success. Figure ~\ref{success-failure-case} illustrates this criterion. Due to the chronic disease long-term  progression where past factors are potentially decisive, all medical histories are included prior to the first diagnosis date. We treat each case as a set of medical records from a patient's EHR as in Figure ~\ref{subset-ehr}. The terms, patient and case, are used interchangeably.

\begin{figure}[htbp]
\centerline{\includegraphics[scale=0.125]{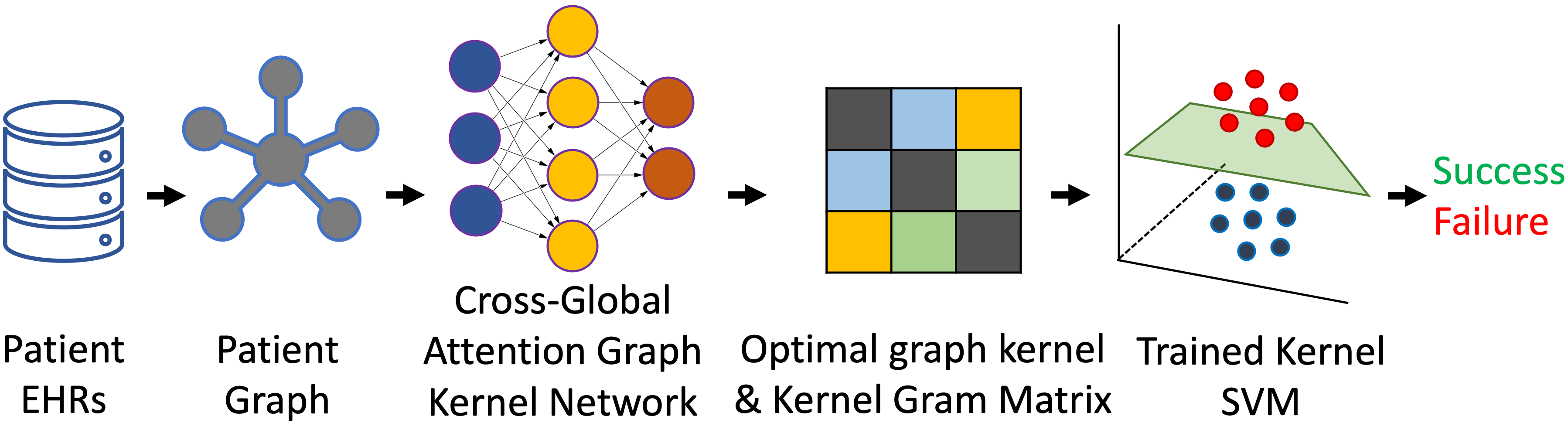}}
\caption{Predictive framework. We create patient graphs to represent all cases. Then, we perform prediction task as a binary graph classification through a kernel SVM. The input kernel gram matrix is generated from Cross-Global Attention Graph Kernel Network.}
\label{framework}
\end{figure}

\begin{figure*}[htbp]
\centerline{\includegraphics[scale=0.27]{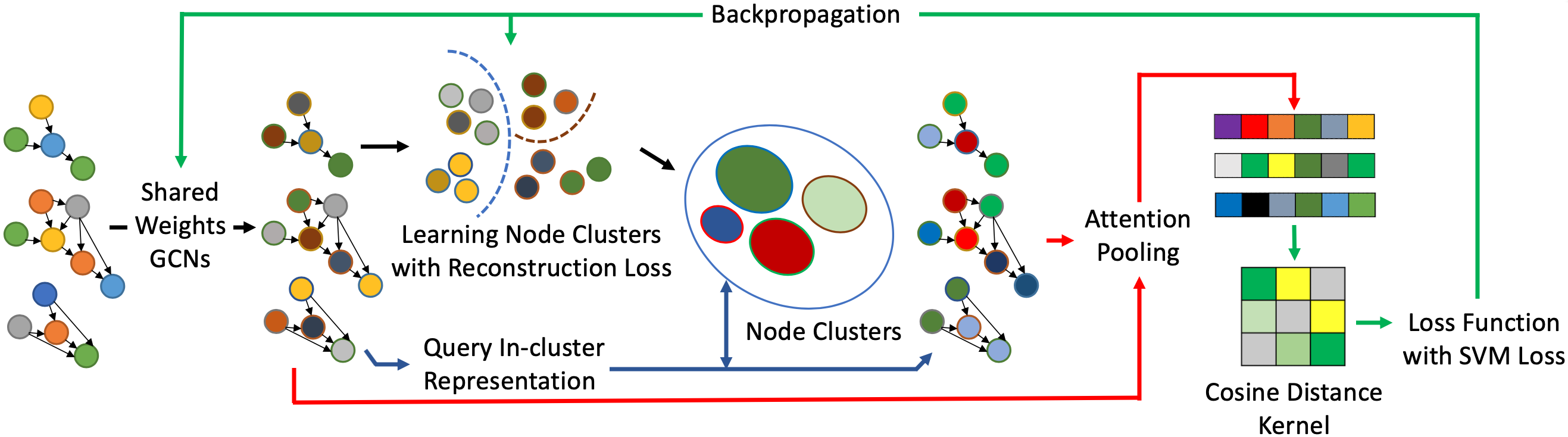}}
\caption{Cross-Global Attention Graph Kernel Network. The node level embedding and node clusters are determined first through black arrows. The graph level embedding is derived (denoted in red arrows) from node matching based pooling (through blue arrows). The loss is calculated by the resulting distance and kernel matrix, and backpropagation is performed to update all model parameters (via green arrows).}
\label{cga-framework}
\end{figure*}

\subsection{Prediction Framework}
We formulate our prediction task as a binary graph classification on graph-based EHR using a kernel SVM. Although similar to ~\cite{yao2019graph}, we differentiate our approach by learning a graph kernel, not handcrafting one. Given a set of success and failure case patient graphs $G$, a deep neural network learns an optimal graph kernel $\mathrm{\textit{{\textbf{k}}}}$. Then, the prediction for success and failure is performed by a kernel SVM using a kernel gram matrix $\mathrm{\textit{{\textbf{K}}}}$ such that $\mathrm{\textit{{\textbf{K}}}}_{ij}=\mathrm{\textit{{\textbf{k}}}}(G_i, G_j)$ where $G_i, G_j \in G$. For an incoming patient, we create a patient graph $G_p$ based on the concatenation of patient's medical history, current diagnosis, and treatment plan. Then, we determine the kernel value between $G_p$ and all training examples $G_i \in G$, and perform prediction through a kernel SVM. The proposed system is illustrated in Figure ~\ref{framework}. 

\section{Cross-Global Attention Graph Kernel Network}
Our Cross-Global Attention Graph Kernel Network learns an end-to-end deep graph kernel on a batch of graphs. This is accomplished through cross-global attention node matching without an explicit pairwise similarity computation. Given a batch $B$ of input graphs $G_1, ..., G_{|B|}$ with batch size $|B|$, we embed their nodes into a lower dimensional space, where node structures and attribute information are encoded in dense vectors. A graph level embedding is then produced by a graph pooling operation on node level embedding via cross-global attention node matching. We calculate the batch-wise cosine distance and generate a kernel gram matrix on the entire batch of resulting graph embedding. Finally, the network loss is computed with contrastive loss, kernel alignment, and SVM primal objective. An overview of cross-global attention graph kernel network is illustrated in Figure~\ref{cga-framework}. The remainder of this section details this process.

\subsection{Graph Embedding}
\subsubsection{Graph Convolutional Networks}
Graph Convolutional Networks (GCN) ~\cite{kipf2016semi} perform 1-hop neighbor feature aggregation for each node in a graph. The resulting graph embedding is permutation invariant when a pooling operation is properly chosen. Given an $n$ number nodes patient graph $G$ with node attribute one-hot vector matrix $X \in R^{n \times c}$, where $c$ denotes the total number of medical codes in EHRs, and a weighted adjacency matrix $A \in R^{n \times n}$, we use GCN to generate a node level embedding $H \in R^{n \times d}$ with embedding size $d$ as follows:

\begin{equation}
H = f({\Tilde{D}}^{-1}\Tilde{A}XW)
\end{equation}

\noindent
where $\Tilde{D}$ is the diagonal node degree matrix of $\Tilde{A}$ defined with $\Tilde{D} = \sum{}_{j} \Tilde{A_{ij}}$, $\Tilde{A} = A + I$ is the adjacency matrix with self-loops added, $W \in R^{c \times d}$ is a trainable weight matrix, and $f$ is a non-linear activation function such as $ReLU(x) = max(0,x)$. The embedding $H$ can be an input to another GCN, creating stacked multiple graph convolution layers:

\begin{equation}
H^{k+1} = f({\Tilde{D}}^{-1}\Tilde{A}H^{k}W^{k}), \quad H^{0} = X
\end{equation}

\noindent
where $H^{k}$ is the node embedding after the $k^{th}$ GCN operation, and $W^{k}$ is the trainable weight associated with the $k^{th}$ GCN layer. The resulting node embedding $H^{k+1}$ contains k-hop neighborhood structure information aggregated by graph convolution layers. 

\subsubsection{Higher-order graph information} 
To capture longer distance nodes and preserve their hierarchical multi-hop neighborhood information as in ~\cite{chen2019dagcn}, we stacked $t$ multiple GCN layers\footnote{Assuming the dimension of all layers' trainable weight matrices are the same} and concatenated all layer's outputs $H^{1:t} = [H^{1},...,H^{t}]$ where $H^{1:t} \in R^{n \times (t \times d)}$. The concatenated node embedding might be very large and could potentially cause a memory issue for subsequent operations. To mitigate such drawbacks, we perform a non-linear transformation on $H^{1:t}$ by a trainable weight $W_{concat} \in R^{(t \times d) \times d}$ and a ReLU activation function as follows:

\begin{equation}
\label{final_embedding}
H_{final} = ReLU(H^{1:t}W_{concat})
\end{equation}

To produce the graph level embedding, instead of using another type of pooling operation ~\cite{zhang2018end, ying2018hierarchical, lee2019self}, we propose cross-global attention node matching and its derived attention based pooling.

\subsection{Cross-Global Attention Node Matching}
Node matching between graphs is computed via a pairwise node similarity measurement. This optimizes a distance metric-based or KL-divergence loss on the graph pairs or triplets~\cite{bai2019simgnn, li2019graph, al2019ddgk} necessitating vast training pairs or triplets to capture the entire global characteristics. One way to avoid explicit pair or triplet generation utilizes efficient batch-wise learning via optimizing classification loss~\cite{wen2016discriminative, qian2019softtriple}. However, 
pairwise node matching in a batch-wise setting is problematic due to graph size variability. 

To address this issue, we propose a novel batch-wise attention-based node matching scheme, a.k.a., cross-global attention node matching. The matching scheme learns a set of global node clusters and computes the attention weight between each node and the representation associated with its membership cluster. The pooling operation based on its attention score to global cluster performs a weighted sum on nodes to derive a single graph embedding.

\subsubsection{Global Node Cluster Learning and Cluster Representation Query} 
Given node embedding $H_{final} \in R^{n \times d}$ from the last GCN layer and transformation after concatenation in Equation~\ref{final_embedding}, we define $M \in R^{s \times d}$ as a trainable global node cluster matrix with $s$ clusters and $d$ dimension features sized to provide an overall representation of its membership nodes. Here, we define membership assignment $A \in R^{n \times s}$ for $H_{final}$ and as follows:

\begin{equation}
    A = Sparsemax(ReLU(H_{final}M^{T}))
\end{equation}

\noindent
where Sparsemax~\cite{martins2016softmax} is a sparse version Softmax, that outputs sparse probabilities. It can be treated as a sparse soft cluster assignment. We can interpret $A$ as a cluster membership identity with $s$ dimension feature representation. We further define the query of nodes' representation in their belonging membership cluster:

\begin{equation}
    Q = Tanh(AM)
\end{equation}

\noindent
where $Q \in R^{n \times d}$ denotes a queried representation for each node in $H_{final}$ from their belonging membership cluster. 

As described in Figure~\ref{example-assignment-matching}, matching can be treated as retrieving cluster identity from global node clusters, and similar nodes are assigned to a similar or even the same cluster membership identity. To construct a better cluster, we add an auxiliary loss by minimizing the reconstruction error, which is similar to Non-negative Matrix Factorization (NMF) clustering in ~\cite{ding2005equivalence} as:
\begin{equation}
    \mathcal{L}_{recon} = {\lVert H_{final} - Q \rVert}_{F}
\end{equation}

\subsubsection{Pooling with Attention-based Node Matching}
\begin{figure}[htbp]
\centerline{\includegraphics[scale=0.14]{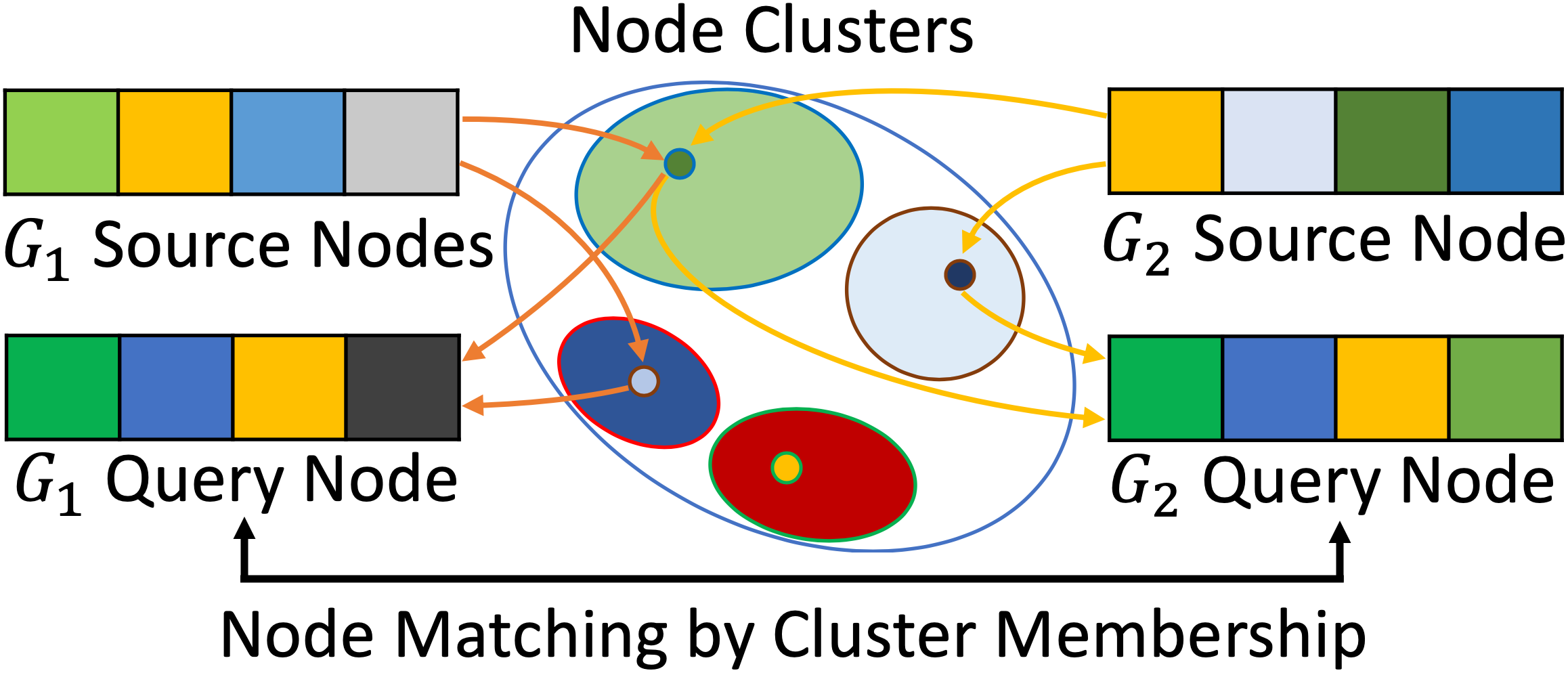}}
\caption{Predictive framework. Each node in $G_1,G_2$ will map to a cluster. Their cluster membership assignments generate their query, which is their representation in terms of belonging cluster. Such an assignment can be seen as a soft label of cluster membership identity. Similar query means similar cluster membership identity, inducing possible matching.}
\label{example-assignment-matching}
\end{figure}

The intuition of pairwise node matching is to assign higher attention-weight to those similar nodes. In other words, matching occurs when two nodes are highly similar, closer to each other than to other possible targets. Following this idea, we observe that two nodes are matched if they have similar or even identical cluster membership. The higher the similar membership identity, the higher the degree of node matching. In addition, a cluster is constructed by minimizing the reconstruction error between the original node $H_{final}$ and the query representation $Q$. A node with high reconstruction error means no specific cluster assignment and further lowers the chance to match other nodes. This can be measured by using entry-wise similarity metrics (e.g., cosine similarity) between ${H_{final}}_{i} \in H_{final}$ and its respective query representation $Q_i \in Q$. Higher similarity between them reveals better reconstruction quality and potential to match other nodes. Based on these observations, we design the cross-global attention node matching pooling, where a node similar to the representation in its cluster membership should receive higher attention weight, as follows:

\begin{equation}
\begin{split}
    \alpha &= Softmax(Sim(H_{final}, Q))  \\
    G_{emb} &= \sum_{i=1}^{n}{\alpha}_{i}{H_{final}}_{i}
\end{split}
\end{equation}

\noindent
where $\alpha \in R^{n}$ is the attention weight for each node, Softmax is applied to generate importance among nodes by using Sim, a similarity metric (e.g., cosine similarity), and the resulting pooling $G_{emb}$ is the weighted sum of node embeddings that compress higher order structure and node matching information from other graphs. Matching and cluster assignment membership is illustrated in Figure ~\ref{example-assignment-matching}.

\subsection{Graph Kernel}
Given a graph pair with their graph level embeddings ${G_{emb}}_{1}, {G_{emb}}_{2}$, we define the graph kernel as follows:
\begin{equation}
    \begin{split} 
        {Dist}_{C}({G_{emb}}_{1}, {G_{emb}}_{2}) &= 1 - \dfrac{\langle{G_{emb}}_{1}, {G_{emb}}_{2}\rangle}{\lVert {G_{emb}}_{1} \rVert \cdot \lVert {G_{emb}}_{2} \rVert} \\
        {Dist}_{E}({G_{emb}}_{1}, {G_{emb}}_{2}) &= {\lVert {G_{emb}}_{1} - {G_{emb}}_{2} \rVert}_{2} \\
        K({G_{emb}}_{1}, {G_{emb}}_{2}) &= exp(-{Dist({G_{emb}}_{1}, {G_{emb}}_{2})}^{2})
    \end{split}
\end{equation}

\noindent
where $Dist_{C}$ is a cosine distance and $Dist_{E}$ is the Euclidean distance. As usual, $\langle \cdot \rangle$ is a standard inner product. $Dist$ can be either $Dist_{C}$ or $Dist_{E}$. The resulting kernel function is positive definite since $exp(-x^2)$ is positive definite for any real number $x$ ~\cite{calin2010heat}\footnote{Due to the space limits, we do not include the complete proof here. It can be easily proved by the definition of positive definiteness with the same derivation in ~\cite{calin2010heat,chang2018bochner}.}. Cosine distance enjoys benefits in more complex data representations. Euclidean distance considers vector magnitude (i.e., norm) during measurement which is not sufficiently sensitive to highly variant features such as long-term disease progressions. Moreover, cosine distance can measure objects on manifolds with nonzero curvature such as spheres or hyperbolic surfaces. In general, Euclidean distance can only be applied to local problems which may not be sufficient to express complex feature characteristics ~\cite{calin2009subriemannian}. The resulting cosine guided kernel is more expressive, and thus, capable of performing implicit high dimensional mapping ~\cite{calin2010heat}.

\subsection{Training}
Given a batch $B$ of input graphs and their class labels $y \in R^{|B| \times 1}$ where $y_{i} \in \{1,0\}$, we get their graph level embeddings for the entire batch via shared weight GCN with cross-global node matching pooling. To support graph size variation within a batch, we concatenate their feature matrices and combine their adjacency matrices into a sparse block-diagonal matrix. Each block corresponds to an adjacency matrix of a graph in the batch. The resulting concatenated feature matrix and block-diagonal matrix are treated as a single graph, so all operations (e.g., GCN and pooling) can perform simultaneously on batch-wise graphs. Then, we calculate their batch-wise distance matrix $\mathrm{\textit{{\textbf{D}}}} \in R^{|B| \times |B|}$ and batch-wise kernel gram matrix $\mathrm{\textit{{\textbf{K}}}} \in R^{|B| \times |B|}$. The model can be trained by mini-batch Stochastic Gradient Descent (SGD) without training pair and triplet generation. To learn an optimal graph embedding, which results in an optimal graph kernel, we optimize it by contrastive loss ~\cite{hadsell2006dimensionality} with a margin threshold $\lambda > 0$:

\begin{equation}
    \mathcal{L}_{contrastive} = \dfrac{1}{|B|}\sum_{i,j \in B}(1 - {\mathrm{\textit{\textbf{Y}}}}_{ij}){max(0, \lambda - {\mathrm{\textit{{\textbf{D}}}}}_{ij})}^{2} + {\mathrm{\textit{\textbf{Y}}}}_{ij}{\mathrm{\textit{{\textbf{D}}}}}_{ij}
\end{equation}

\noindent
and kernel alignment loss~\cite{cristianini2002kernel}:

\begin{equation}
    \mathcal{L}_{alignment} = \dfrac{1}{|B|}\sqrt{2 - 2( {\langle\mathrm{\textit{{\textbf{K}}}},\mathrm{\textit{\textbf{Y}}}  \rangle}_{F}\mathbin{/}\sqrt{{\langle\mathrm{\textit{{\textbf{K}}}},\mathrm{\textit{\textbf{K}}}  \rangle}_{F}{\langle\mathrm{\textit{{\textbf{Y}}}},\mathrm{\textit{\textbf{Y}}}\rangle}_{F}})}
\end{equation}

\noindent
where ${\langle\cdot\rangle}_{F}$ denotes the frobenius inner product, $\mathrm{\textit{\textbf{K}}}$ is a batch-wise kernel gram matrix, and $\mathrm{\textit{\textbf{Y}}} \in R^{|B| \times |B|}$ where $\mathrm{\textit{\textbf{Y}}}_{ij} = 1$ if $y_{i} = y_{j}$ else $\mathrm{\textit{\textbf{Y}}}_{ij} = 0$. We believe that a good distance-metric induces a good kernel function and vice versa. So, we learn the graph kernel jointly through optimal cosine distance between graphs via contrastive loss with an optimal graph kernel through kernel alignment loss.

To align a learned embedding, distance, and kernel to the classification loss in  end-to-end training, we incorporate the SVM primal ~\cite{chapelle2007training} objective with squared hinge loss function into our objective:

\begin{equation}
    \begin{split}
       \mathcal{L}_{SVM} = \frac{1}{C}\sum_{i,j \in B}\beta_{i}\beta{j}{\mathrm{\textit{{\textbf{K}}}}}_{ij} + 
        \sum_{i}{max(0, 1 - y_{i}\sum_{j \in B}{\mathrm{\textit{{\textbf{K}}}}}_{ij}\beta_{j})}^{2}
    \end{split}
\end{equation}

\noindent
where $C >= 0$ is a user defined regularization constant and $\beta \in R^{|B| \times 1}$ is a trainable coefficient weight vector. The following is the final model optimization problem formulation:

\begin{equation}
\begin{aligned}
\min_{\theta, \beta} \quad (\underset{\theta}{\mathcal{L}_{contrastive}} +  \underset{\theta}{\mathcal{L}_{alignment}} + \underset{{\theta}}{\mathcal{L}_{recon}}) + \underset{{\beta}}{\mathcal{L}_{SVM}}
\end{aligned}
\end{equation}

\noindent
where $\theta$ denotes a set of all trainable variables in graph embedding and $\beta$ is a trainable coefficient weight vector for SVM. Since the training is done by mini-batch SGD, the SVM objective is only meaningful for a given batch. Namely, gradient for $\beta$ in SVM are only relevant for the current batch update as the SVM objective is dependent on the input kernel gram matrix. When training proceeds to the next batch, the kernel gram matrix is different, and the optimized $\beta$ is inconsistent with the last batch status. To resolve this inconsistent weight update problem, we treat SVM as a light-weight auxiliary objective (e.g., regularization),  encouraging the model to learn an effective graph kernel. In this case, we first perform a forward pass through graph kernel network, then we train the SVM by feeding in the kernel gram matrix from the forward pass output until convergence. The positive definiteness of the kernel function guarantees SVM convergence\footnote{Due to page limit, we do not show convergence analysis. However, averaged iterations for SVM is 18.}. Once the SVM is trained, we treat $\beta$ as a model constant, and $\mathcal{L}_{SVM}$ now acts as a regular loss function. The gradient of $\theta$ can be computed through ${\mathcal{L}_{kernel}}$, ${\mathcal{L}_{recon}}$, and ${\mathcal{L}_{SVM}}$, and the model can perform backpropagation to update $\theta$.

\section{Experiments}
\subsection{Dataset}
We evaluate our model on real-world EHRs, a subset of the Taiwanese National Health Insurance Research Database (NHIRD) \footnote{https://nhird.nhri.org.tw/en/}, which contains over a 20-year complete medical history for one-million randomly sampled de-identified patients. NHIRD composes reimbursement related registration files and original claim data for hospitals and clinics that enroll in the National Health Insurance (NHI) program. The ICD9-CM\footnote{International Classification of Diseases, 9th Revision, Clinical Modification} code indicates the diagnosed disease and the ATC\footnote{Anatomical Therapeutic Chemical} code is used for drug prescription. Institutional Review Board (IRB) approvals for our research were granted by all associated institutions.

The three most prevalent chronic diseases in Taiwan, namely, hypertension, hyperlipidemia, and diabetes, are selected. Their treatments primarily rely on a long-term treatment plan including multiple drug prescriptions to control disease progression. The effectiveness of treatment depends on the risk level of possible future severe comorbidities and complications after receiving the treatment plan for several years. The goal is to predict the success or failure for the given drug prescriptions during the treatment period of a chronic disease diagnosis. According to our collaborating medical doctors and to published treatment guidelines of hypertension ~\cite{chiang20152015}, hyperlipidemia~\cite{li20172017}, and diabetes~\cite{diabetes-guideline}, we define success and failure cases for each disease by the following steps:

\begin{enumerate}
    \item Locate the first chronic disease diagnosis date $T_0$. 
    \item Set ${Y}_{plan}$ year observation window for treatment plan.
    \item Set treatment plan end date $T_{plan} = T_0 + {Y}_{plan}$.
    \item Set ${Y}_{outcome}$ year observation window for outcome.
    \item Set outcome observation end date $T_{outcome} = T_{plan} + {Y}_{outcome}$.
    \item If no selected severe comorbidities and complications diagnosis exist between $T_{plan}$ and $T_{outcome}$, the case is defined successful, otherwise, a failure.
\end{enumerate}

\noindent
We use the patient's entire medical history (a.k.a., clinical visits) from the first medical record to $T_{plan}$ to create the patient graphs. We set ${Y}_{plan}$ for 1 year and ${Y}_{outcome}$ for 10 years. For each medical event, we extract all diagnosis ICD-9 codes and drug prescription ATC codes. Table \ref{Disease-Stat} summarizes the dataset statistic, and Table \ref{complication_table} lists all selected complication ICD-9 codes for each disease.

\begin{table}[]
\caption{Selected Complication ICD9 Codes}
\label{complication_table}
\begin{tabular}{cl}
\hline
Disease        & \multicolumn{1}{c}{Selected Complication ICD9 Codes}                                                                                                                                                                                                                                                                                                                                                                      \\ \hline
Hypertension   & \begin{tabular}[c]{@{}l@{}}402.*: Hypertensive heart disease                  \\ 403.*: Hypertensive renal disease    \\ 404.*: Hypertensive heart and renal disease  \\ 410.*: Acute myocardial infarction\\ 428.*: Heart failure \\ 434.*: Occlusion of cerebral arteries\end{tabular}                                                                                       \\ \hline
Hyperlipidemia & \begin{tabular}[c]{@{}l@{}}410.*: Acute myocardial infarction\\ 411.*: Other acute and subacute \\ \qquad \enspace forms of ischemic heart disease\\ 412.*: Old myocardial infarction \\ 413.*: Angina pectoris of heart disease\\ 43*.*: Cerebrovascular disease\end{tabular} \\ \hline
Diabetes       & \begin{tabular}[c]{@{}l@{}}361.*: Retinal detachments and defects\\ 362.*: Other retinal disorders\\ 365.*: Disorders of iris and ciliary body\\ 366.*: Cataract\\ 369.*: Blindness and low vision\end{tabular}                                                                                                                                                                                                           \\ \hline
\end{tabular}
\end{table}

\begin{table}[]
\caption{Dataset Statistics. The percentage denotes data imbalance ratio especially in hyperlipidemia and diabetes.}
\label{Disease-Stat}
\begin{tabular}{cccc}
\hline
Disease             & Hypertension   & Hyperlipidemia & Diabetes      \\ \hline
ICD9 Codes          & 401.*          & 272.*          & 250.*         \\ \hline
\# of patient       & 235,695        & 123,380        & 131,997       \\
\# of failure       & 104,936 (45\%) & 26,043 (21\%)  & 34,414 (26\%) \\
\# of success       & 130,759 (55\%) & 97,337 (79\%)  & 97,583 (74\%) \\ \hline
Max \# nodes        & 33,497         & 19.159         & 15,454             \\
Min \# nodes        & 3              & 3              & 3             \\
Avg \# nodes        & 220            & 285            & 374           \\ \hline
Max \# edges        & 87,852         & 52,750         & 57422             \\
Min \# edges        & 2              & 2              & 2             \\
Avg \# edges        & 561            & 620            & 891           \\ \hline
\end{tabular}
\end{table}

\begin{table*}[]
\caption{Performance comparison. We can see our proposed model outperforms all baselines especially for imbalance disease cases. The superiority of cosine distance over the Euclidean counterpart is apparent on those imbalance cases.}
\label{tab:performance-comparison}
\begin{tabular}{cccccccccc}
\hline
                       & \multicolumn{3}{c}{Hypertension}                    & \multicolumn{3}{c}{Hyperlipidemia}                  & \multicolumn{3}{c}{Diabetes}                        \\ \hline
Model                  & ACC        & AUC             & F1              & ACC        & AUC             & F1              & ACC        & AUC             & F1              \\ \hline
CGA-GK-Cosine (Our)      & \textbf{0.7417} & \textbf{0.7361} & \textbf{0.7371} & \textbf{0.8702} & \textbf{0.7428} & \textbf{0.7727} & \textbf{0.7804} & \textbf{0.6602} & \textbf{0.6758} \\ 
CGA-GK-Euclidean (Our)   & 0.7337 & 0.7278 & 0.7290 & 0.8507 & 0.6798 & 0.7153 & 0.7613 & 0.5920 & 0.5970 \\ \hline
MGKF                   & 0.6990          & 0.7025          & 0.6973          & 0.7200          & 0.7043          & 0.6698          & 0.7250          & 0.6354          & 0.6404          \\
WL-Kernel-SVM          & 0.7101          & 0.6968          & 0.6982          & 0.8293          & 0.6092          & 0.6304          & 0.7625          & 0.5911          & 0.5955          \\
DGCNN                  & 0.6954          & 0.6895          & 0.6894          & 0.8290          & 0.6338          & 0.6518          & 0.7536          & 0.5871          & 0.5914          \\ \hline
ClinicalBERT           & 0.7132          & 0.6996          & 0.6434          & 0.8510          & 0.6808          & 0.5215          & 0.7720          & 0.6484          & 0.4718          \\
Retain                 & 0.6580          & 0.6537          & 0.6174          & 0.8340          & 0.6908          & 0.5337          & 0.7657          & 0.6369          & 0.4529          \\
Dipole                 & 0.6603          & 0.6805          & 0.6782          & 0.8180          & 0.5943          & 0.3259          & 0.7553          & 0.5540          & 0.2338          \\
LSTM                   & 0.6960          & 0.6607          & 0.5250          & 0.7920          & 0.6267          & 0.3988          & 0.7283          & 0.5497          & 0.2598          \\
CNN                    & 0.7170          & 0.6999          & 0.6323          & 0.8320          & 0.6920          & 0.5359          & 0.7317          & 0.6481          & 0.4679          \\
Med2Vec                & 0.6864          & 0.6681          & 0.5847          & 0.8167          & 0.6593          & 0.4575          & 0.7524          & 0.2698          & 0.5805          \\
Deep Patient           & 0.6560          & 0.6443          & 0.5835          & 0.7980          & 0.5295          & 0.1217          & 0.7280          & 0.5395          & 0.1905          \\ \hline
LR       & 0.7220          & 0.7083          & 0.6532          & 0.8368          & 0.6417          & 0.4401          & 0.7483          & 0.5371          & 0.1658          \\
SVM      & 0.6909          & 0.6738          & 0.6016          & 0.8168          & 0.6939          & 0.5252          & 0.7294          & 0.6329          & 0.4544          \\
RF      & 0.7266          & 0.7188          & 0.6774          & 0.8424          & 0.6507          & 0.4609          & 0.7631          & 0.3066          & 0.5811          \\ \hline
\end{tabular}
\label{Evaluation-result}
\end{table*}

\subsection{Baselines}
Three types of baselines are selected to compare our model performance: Deep learning based, graph classification based, and traditional based.

\textbf{Deep learning based approaches}: 
\begin{itemize}
    \item Deep Patient ~\cite{deep-patient}. Deep Patient utilizes a three-layer stacked denoising autoencoder to perform unsupervised representation learning on EHRs with Random Forest to predict future diagnosis. 
    \item LSTM ~\cite{lstm}. A LSTM model with word embedding to encode time series clinical measurements in EHRs is used to predict future medical code diagnosis.
    \item Med2Vec ~\cite{med2vec}. Med2Vec uses multi-layer perceptron to learn interpretable code and visit embedding based on the skip-gram model. The code level embedding is learned first, and the resulting embedding is concatenated with demographic information to form visit level embedding.
    \item Retain ~\cite{retain}. It is a RNN using GRU with a two-level reverse time attention mechanism, which offer interpretation to select influential past visits contributing to the final prediction.
    \item CNN ~\cite{CNN}. This model uses a 1D-CNN to learn EHRs temporal embedding matrix representation to capture local and short temporal dependency in EHRs for risk prediction.
    \item Dipole ~\cite{dipole}. It is a bidirectional RNN with three different attention mechanisms, proposed to calculate attention weights for each patient visit: general, concatenation-based, and location-based. In our experiment, we use multiplicative attention ~\cite{luong2015effective} to compute attention weight due to memory constraints.
    \item Clinical BERT ~\cite{clinicalBert}. In their work, a pre-trained clinical language model trained by the state-of-the art BERT model is created. We use their Clinical BERT as the BERT base model to train our language model on NHIRD. For each patient case, we concatenate all medical codes from all visits into a single document. Then, we fine-tune it on our prediction task.
\end{itemize} 

\textbf{Graph based approaches}:
\begin{itemize}
    \item WL-Kernel-SVM ~\cite{shervashidze2011weisfeiler}. Here, we use Weisfeiler-Lehman subtree graph kernel to compute a pairwise kernel gram matrix on all patient graphs. Then, a kernel SVM is used to perform graph classification.
    \item DGCNN ~\cite{zhang2018end}. It is an end-to-end graph classification model by graph convolution networks with a sort pooling layer to derive permutation invariant graph embeddings. 1D-CNN then extracts features along with full-connected layer for graph classification task on patient graphs.
    \item MGKF ~\cite{yao2019multiple}. In their work, a deep learning architecture to learn the fusion representation of three types of graph kernels is proposed. They perform an antibiotics-based disease drug prediction task. In our experiment, we replace their shortest path kernel with Weisfeiler-Lehman subtree graph kernels on patient graphs to avoid insufficient memory and a forever running time issue for shortest path kernel.
\end{itemize}

\textbf{Traditional approaches~\footnote{All patient cases are represented as documents with one-hot encoding containing all medical codes from all visits.}}:

\begin{itemize}
    \item Linear Support Vector Machine (SVM).
    \item Logistic Regression (LR).
    \item Random Forest (RF).
\end{itemize}

\subsection{Evaluation Setup}
\label{evaluation-setup}
Accuracy (ACC), F1-score (Macro F1), and the area under the receiver operating characteristic curve (AUROC) are used as our evaluation metrics. For each disease, we randomly divide our datasets into training, validation, and testing sets in an 80:10:10 ratio. We notice the data imbalance as shown in Table ~\ref{Disease-Stat}. To reflect real-world clinical practice, we do not use any data balancing techniques and keep data imbalance. All parameters for all evaluated models are fine tuned via the validation set. The pairwise t-test with a p-value set to 0.05 is used to reject the null-hypothesis to assess the statistical significance of our proposed model. Our solution statistically significantly differs from previous efforts. For our proposed model architecture, we set 6 layers GCN with output dimension 256 and ReLU activation function. We set the number of global node clusters to 256 and the contrastive loss margin threshold $\lambda=1$. We use Tensorflow-Keras to implement our proposed model architecture. For SVM training, we set 100 iterations with early stopping and an regularization constant $C=1$. For the training stage, we use the Adam optimizer with an initial fixed learning rate set to 0.0005 with 128 batch size and train for 10 epochs with early stopping criteria. For the graph classification stage, we use classical kernel SVM from scikit-learn ~\cite{scikit-learn} and set the regularization constant $C=1$. All experiments are executed on an Intel Core i7 CPU, with 64GB memory and one Nvidia 1080 Ti GPU.

\subsection{Experimental Results} \label{results}
Table \ref{Evaluation-result} shows that our proposed approach (CGA-GK-Cosine) consistently outperforms all baseline approaches on all evaluation metrics. Specifically, all baselines are affected by data imbalance and receive high Accuracy and AUC but low F1 scores, particularly in the imbalance hyperlipidemia and diabetes dataset as depicted in Table ~\ref{Disease-Stat}. Data imbalance is common in real-world clinical practice, and it is critical when developing medical applications. It is undesirable to prescribe a false predicted success drug treatment, which may lead to severe disease progression or fatality. Furthermore, NHIRD, a real-world claim-based EHR database, is known to have highly biased medical records along with unpredictable and irregular patterns such as (1) Record splitting: multiple same diagnosis records with different drug prescription (2) Reimbursement trick: only record higher reimbursement drug or disease (3) Patient shopping behavior: multiple same disease diagnosis without drug prescription on the same date, and medical events from all of these conditions are pointless. 

Our approach is insensitive to data imbalance and yields the highest F1 score, highlighting its ability to learn a meaningful and noise resistant graph kernel since the prediction is conducted purely by traditional kernel SVM. The high F1 score also demonstrates that CGA-GK-Cosine outperforms  CGA-GK-Euclidean, highlighting the advantage of cosine distance over Euclidean distance. This result confirms our hypothesis that cosine distance captures micro differences in feature dimension and is relatively insensitive to the highly biased dataset, as compared to its counterparts.

Looking at different baseline groups, the graph-based approach outperforms all other baselines on F1, revealing the usefulness for graphs as a modeling tool under real world data imbalance situation. The graph kernel approaches show the effectiveness of similarity-based classification to overcome highly variant and imbalanced medical records. For deep learning baselines, we observe they all tend to predict drug treatment as success, which leads to low F1 for all tasks. It is even worse on RNN based models due to their over-fitting on the majority class. We also hypothesize that pre-trained fine-tuned BERT language model is not suitable for drug prediction task, as its training objective is not aligned to disease progression. 

For most other research efforts, the datasets used are either from a collaborating hospital or public dataset, namely MIMIC3\footnote{We're not using MIMIC3 since it doesn't contain enough medical history to monitor chronic disease outcome.}, with a significantly shorter medical history per patient and much less biased data records. Consequently, the model developed on such datasets fails to comply with NHIRD and suffers from over-fitting under an imbalanced situation. Traditional approaches are too shallow to learn meaningful representation; however, due to their simple learning process, they can avoid severe over-fitting and perform better than some deep learning approaches (e.g., Dipole, LSTM, and Deep Patient). 

\subsection{Cross-Global Attention Node Matching}
\begin{figure}
\centering
\vfill
\subfigure[All node labels and edge connections are removed. \label{matching-result-1}]{\includegraphics[scale=0.13]{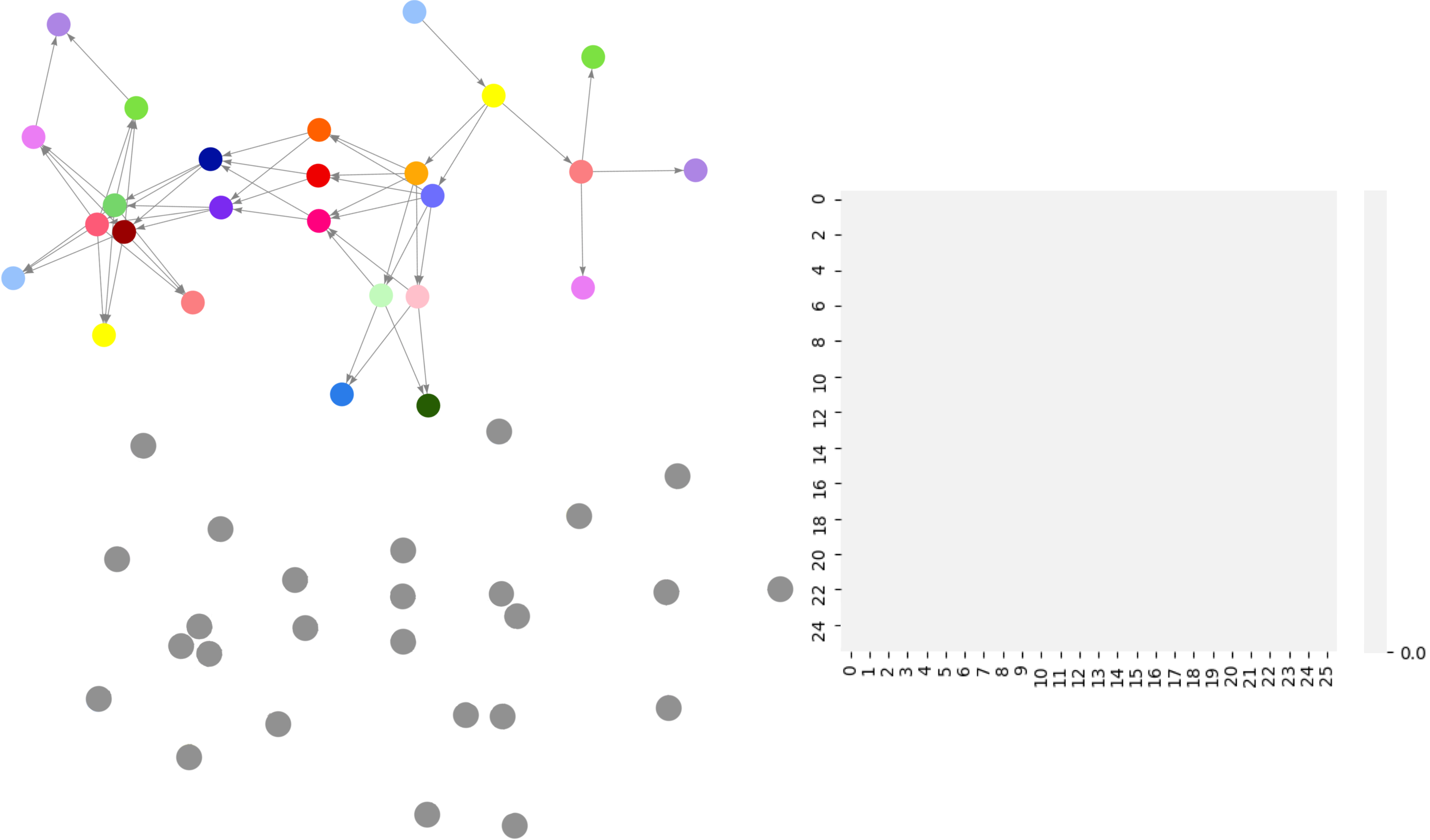}}
\vfill
\subfigure[Recover some node labels and all edge connections. \label{matching-result-2}]{\includegraphics[scale=0.13]{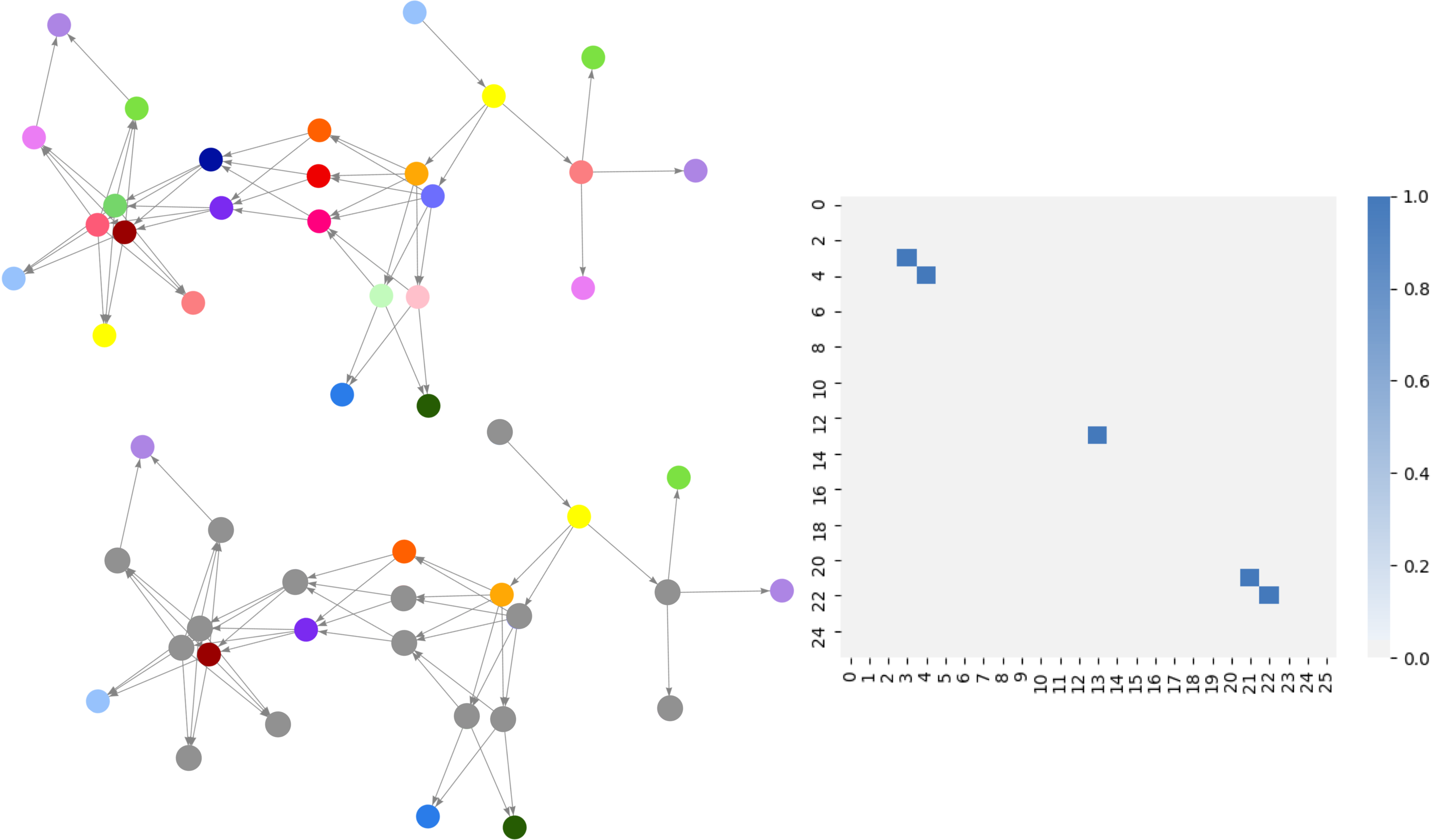}}
\vfill
\subfigure[Recover all node labels. \label{matching-result-3}]{\includegraphics[scale=0.13]{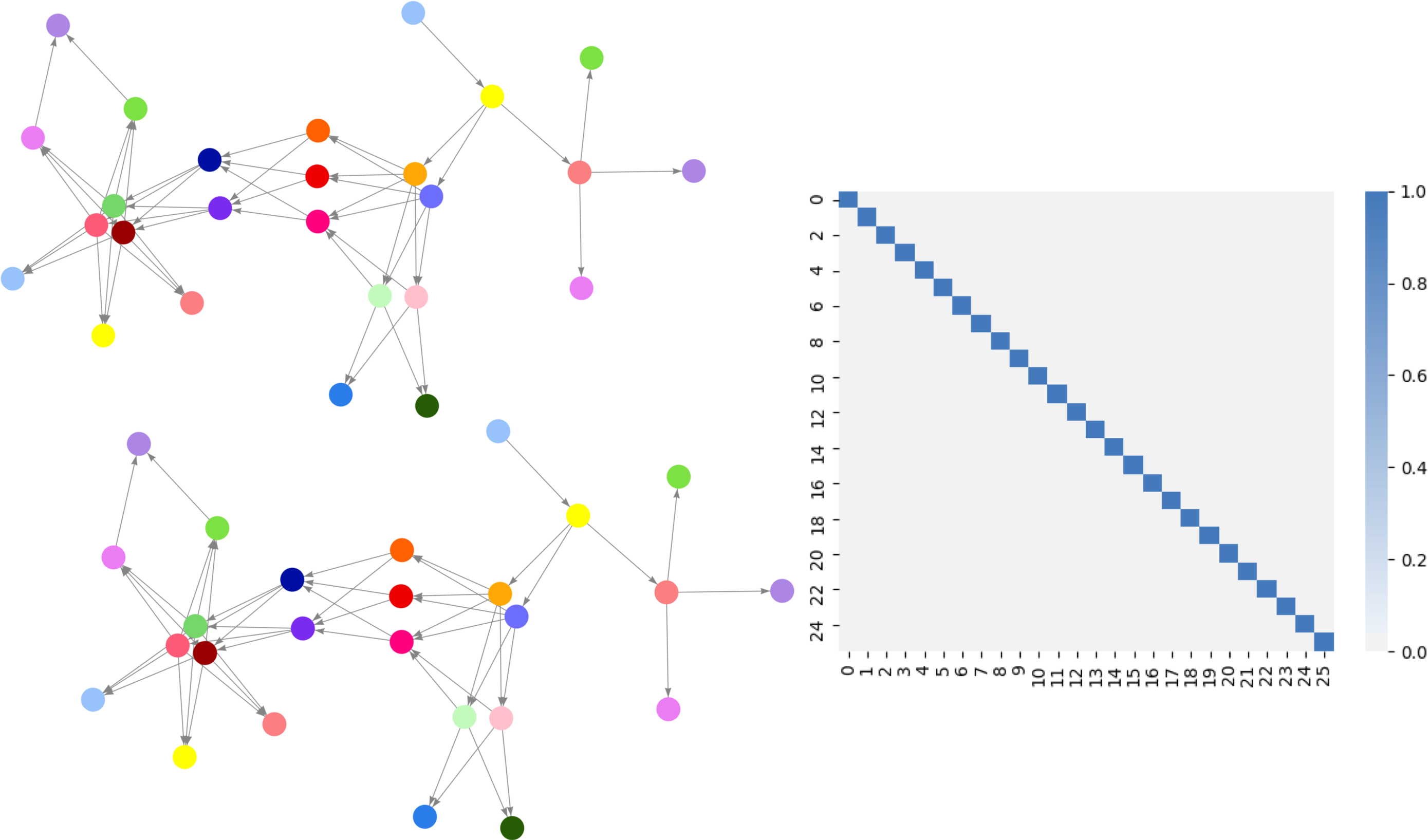}}
\caption{Cross-global Attention Node Matching on hypertension success case patient graph. Diagonal means self matching. We can see how self-matching changes when we recover some node labels and edges.}
\label{matching-result}
\end{figure}

We evaluate our proposed node cluster membership assignment based on how two identical graphs relate to each other under different node and edge removals. If two graphs are identical, they should match themselves symmetrically (e.g., the diagonal part). By randomly removing some node and edge labels, the matching result will change since the graph structure is changed. We select an identical patient graph from hypertension in Figure ~\ref{matching-result}. Nodes of the same color indicate same cluster membership by selecting the largest dimension in their cluster membership label, which implies matching. The heatmap shows the full attention matrix on node matching which reveals a more complete view of their alignment. 

In Figure ~\ref{matching-result-1}, when we remove all node labels and edge connections, nodes do not match themselves. At this time, they are considered as differing graphs with different cluster membership assignments, although they are actually identical. By recovering some node labels and all edge connections in Figure ~\ref{matching-result-2}, we can see their degree of alignment is increased. Finally, when all nodes and edges are recovered, their nodes are matched to themselves with the same cluster membership assignment. In Figure ~\ref{matching-result-3}, we can see the matching is a symmetrical one-to-one alignment between all nodes in the diagonal. Results suggest that our cross-global attention node matching, which is computed on batch of graphs simultaneously, can successfully provide an effective matching without explicit pairwise graph comparisons.

\subsection{Model Interpretation}
\label{interpreting-results}
Our proposed method enjoys three types of interpretations: (1) patient graph interpretation, (2) most similar case on cause of prediction, and (3) knowledge discovery on support vectors: \\

\begin{figure}
\centering
\includegraphics[scale=0.14]{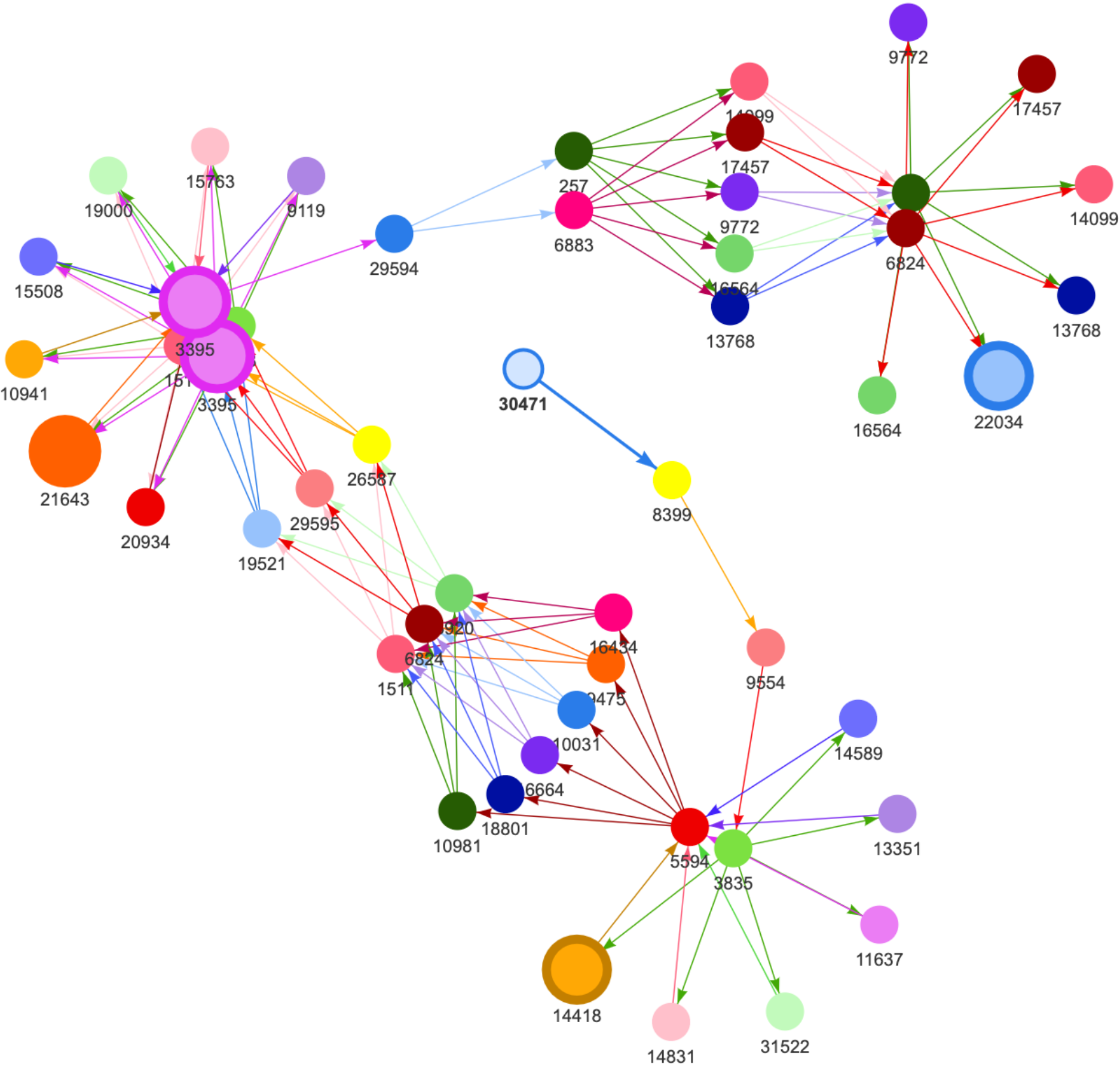}
\caption{An example hyperlipidemia patient graph. Each number denotes a medical code in NHIRD. The directed edge and its length tell disease progression. The size of the node indicates its importance in the patient graph.}
\label{hyperlipidemis-patient-graph}
\end{figure}

\noindent
\textbf{Patient Graph Interpretation} \\
We can use the cross-global attention score on each node to discover important disease diagnoses and drug prescriptions on a per patient basis. The higher the score, namely the better matching to others, the more important the node is in the similarity computation. The patient graph in Figure ~\ref{hyperlipidemis-patient-graph} is easily understood by medical doctors due to their graphical representation. Together with high attentive node visualization, they provide investigative direction and background knowledge on patient disease progression. \\

\begin{figure*}[h]       
    \hfill{\includegraphics[scale=0.24]{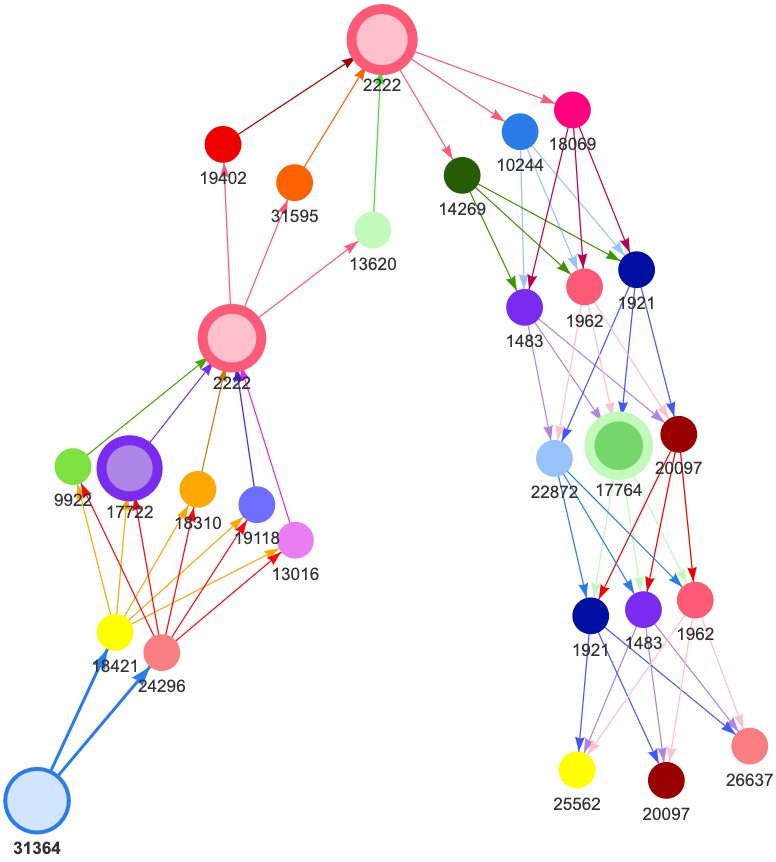}}   
    \hspace{0px}
    \hfill{\includegraphics[scale=0.24]{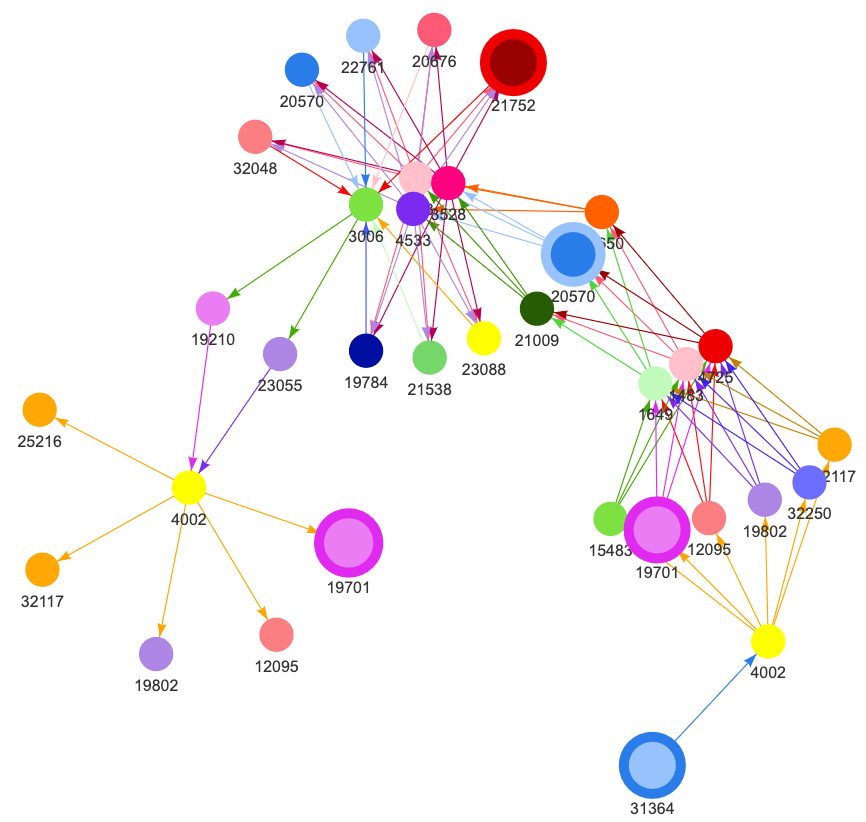}}
    \hspace{0px}
    \hfill{\includegraphics[scale=0.34]{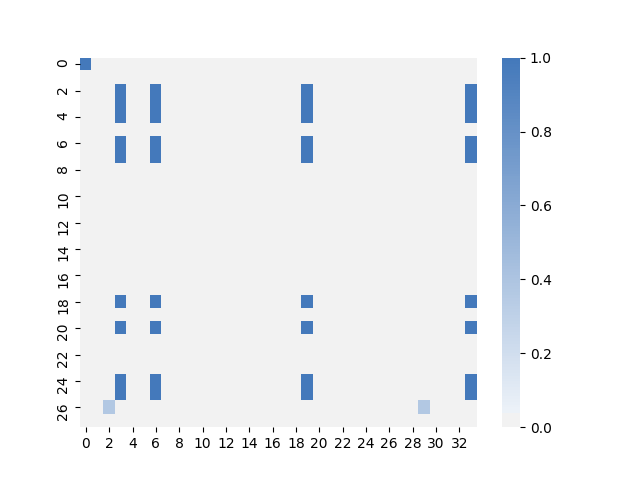}}
    \caption{An example diabetes patient failure case patient graph (left) with the patient's most similar patient graph (middle) and their graph node matching (right). The heatmap of their matching explains what makes these two patient graphs similar.}
    \label{materialflowChart}
    \label{example-matching}
\end{figure*}

\noindent
\textbf{Most Similar Case on Cause of Prediction} \\
Kernel $K(G_i,G_j)$ measures the similarity between two cases patient graph $G_i, G_j$; we can infer the most similar case $G_j$ for $G_i$ by finding the highest kernel value. With cross-global attention node matching in Figure ~\ref{example-matching}, one sees how these two graphs match each other. The graphical representation highlights common disease progression related to matched nodes. The insights on how these two patient cases are similar guides medical doctors as to the cause of why the given treatment is a success or failure. \\

\noindent
\textbf{Knowledge Discovery on Support Vectors} \\
Finally, we consult a set of top support vectors from the kernel SVM, which receive top maximum dual coefficients ~\footnote{Refer ~\cite{chapelle2007training} for SVM dual formulation.}, interpreting the overall importance to assign a class label during SVM training. Combining this with the previous two types of interpretation techniques, we are able to discover knowledge among overall disease patterns for a success or failure treatment plan.

\section{Conclusion and Future Work}
The highly biased and variant nature of real-world EHR coupled with the long-term disease progression behavior of chronic diseases challenge the development for predictive models in clinical decision support. Many proposed prior efforts address such difficulty, yet none succeeded nor earned clinical deployment. Deep learning models tend to over-fit on real-world EHR with highly biased long-term time progression medical patterns, worse yet when data imbalance exists.  Furthermore, interpretability measures still demand refinement due to the opaqueness of deep neural networks. 

Accordingly, we proposed a deep learning model, namely, cross-global attention graph kernel network, to learn an optimal graph kernel and achieve state-of-the-art prediction accuracy on highly biased and imbalanced real-world EHR. The cosine distance guided the learning process with SVM primal objective learning an optimal noise resistant graph kernel. The novel cross-global attention node matching efficiently captures important graph structure without explicit pairwise comparisons. The classification results outperform all state-of-the-art baselines simply using a traditional kernel SVM. Three types of interpretation techniques can work cooperatively to maximize model interpretability. We also notice that cosine distance has interesting properties, specifically, in sub-Riemannian geometry. This is a very active research direction in partial differential equations with many applications in control theory, such as self-driving automobiles and the stochastic process of heat flows~\cite{calin2009subriemannian, calin2010heat}. We plan to study the sub-Riemannian geometry that can be applied to model EHR patient similarity.

Our approach predicts chronic disease drug prescription outcome for long-term disease progression, exceeding 
the performance of state-of-the-art in all evaluation metrics, while providing interpretability. It was intentionally designed in coordination with and is under current use and assessment by medical clinicians in diverse clinical practices. 

\bibliographystyle{ACM-Reference-Format}
\bibliography{}

\end{document}